\documentclass[10pt,twocolumn,letterpaper]{article}

\usepackage{iccv}
\usepackage{times}
\usepackage{epsfig}
\usepackage{graphicx}
\usepackage{amsmath}
\usepackage{amssymb}

\usepackage[breaklinks=true,bookmarks=false]{hyperref}

\iccvfinalcopy 

\ificcvfinal\fi

\begin{document}

\title{Learning to ``Segment Anything'' in Thermal Infrared Images through Knowledge Distillation with a Large Scale Dataset SATIR}

\author{Junzhang Chen\\
Beihang University\\
{\tt\small chenjz@buaa.edu.cn}
\and
Xiangzhi Bai\\
Beihang University\\
{\tt\small jackybxz@buaa.edu.cn}
}

\maketitle
\ificcvfinal\thispagestyle{empty}\fi

\begin{abstract}
The Segment Anything Model (SAM) is a promptable segmentation model recently introduced by Meta AI that has demonstrated its prowess across various fields beyond just image segmentation. SAM can accurately segment images across diverse fields, and generating various masks. We discovered that this ability of SAM can be leveraged to pretrain models for specific fields. Accordingly, we have proposed a framework that utilizes SAM to generate pseudo labels for pretraining thermal infrared image segmentation tasks. Our proposed framework can effectively improve the accuracy of segmentation results of specific categories beyond the SOTA ImageNet pretrained model. Our framework presents a novel approach to collaborate with models trained with large data like SAM to address problems in special fields. Also, we generated a large scale thermal infrared segmentation dataset used for pretaining, which contains over 100,000 images with pixel-annotation labels. This approach offers an effective solution for working with large models in special fields where label annotation is challenging. Our code is available at https://github.com/chenjzBUAA/SATIR
\end{abstract}

\section{Introduction}

		\begin{figure}[htbp]
		\centering
		\includegraphics[width=3in]{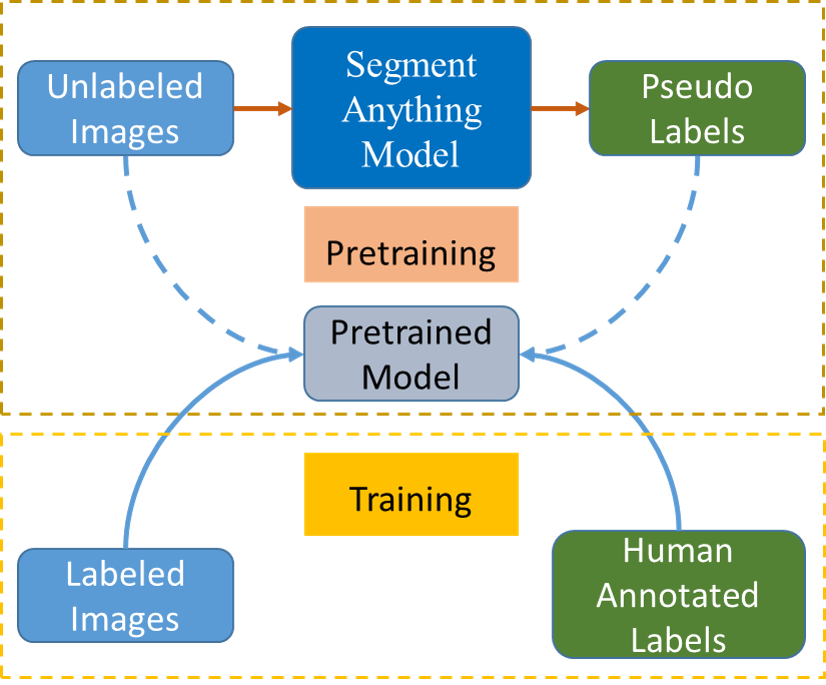}
		\caption{Overview of our proposed method for pretraining thermal infrared image segmentation using knowledge distillation from the Segment Anything Model (SAM) to generate pseudo labels.}
		\label{index}
	\end{figure}
The Segment Anything Model (SAM) has recently been introduced by Meta AI as a promptable segmentation model \cite{kirillov2023segment}, showcasing its effectiveness across diverse fields beyond image segmentation.

SAM has shown promising results in image segmentation tasks, but it may not always provide precise segmentation for thermal images. This imprecision may not be suitable for supporting annotations, as it may require additional time to finetune the annotations by human annotators. However, the coarse segmentation results obtained from SAM, trained on a large-scale dataset, can still be valuable as a pretraining step or as a teacher model to train a segmentation model for specific species segmentation tasks.

Therefore, a knowledge distillation approach is required, leveraging models trained on large-scale datasets like SAM.

The knowledge distillation approach can be categorized into three types \cite{gou2021knowledge}: Feature Based \cite{bengio2013representation}, Response Based \cite{chen2017learning} and Relation Based Knowledge Distillation \cite{yim2017gift}.

\begin{figure*}[htbp]
	\centering
	\includegraphics[width=7in]{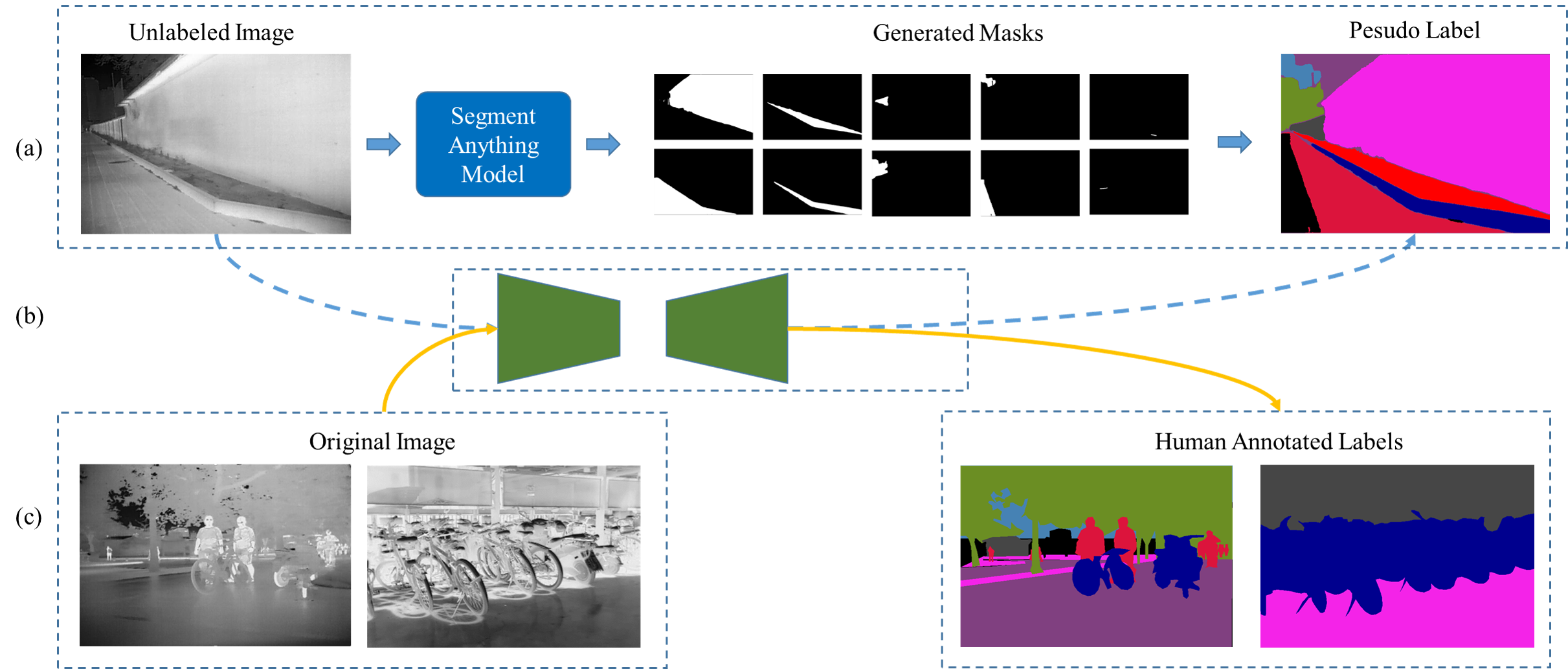}
	\caption{The proposed framework consists of three steps: (a) construction of a pretrained dataset using SAM, (b) pretraining of the model with the dataset, and (c) finetuning the pretrained model on the target task.}
	\label{framework}
\end{figure*}

Here, the domain knowledge required for thermal infrared images differs significantly from that of the images typically used to train SAM. As a result, response-based knowledge distillation is preferred for our task. In our approach, we form a framework which utilize the segment ability of SAM, mainly composed of leveraging the masks generated by SAM to construct pseudo labels for pretraining our model. Accordingly, we build a large scale of pretraining dataset for thermal infrared segmentation task. The dataset contains over 100,000 images with pixel-level annotations generated by our method.

The main contributions of our method are two-fold:

\begin{itemize}
\item We developed a framework for pretraining the model of segmentation tasks in special areas, such as thermal infrared images, by leveraging models trained on large-scale datasets like SAM.
\item We have constructed a large-scale thermal infrared dataset SATIR for segmentation with pixel-level annotations, which can be utilized for pretraining thermal infrared image segmentation tasks. The dataset comprises over 100,000 annotated images, including diverse scenes such as urban, indoor/outdoor, aerial, and more.
\end{itemize}

\section{Related work}
Feature Based Knowledge Distillation \cite{romero2014fitnets} involves learning the parameters in the teacher model to distinguish between features of different objects or classes, by training against a loss function. Each class has a set of parameters (features) that aid in predicting that object or class. During student model training, the same sets of feature activations are learned through distillation loss. The loss function reduces the difference in activation of specific features between the teacher and student models.

Response Based Knowledge Distillation \cite{zhang2019fast,meng2019conditional} is distinct from Feature Based Knowledge Distillation as it involves the student model utilizing the teacher model's logits to enhance its own. The distillation loss in this approach is focused on minimizing the variance in logits between the two models. The student model mimics the teacher model's predictions during the training process to improve its performance.

Relation Based Knowledge Distillation \cite{lee2018self} differs from Feature and Response-Based Knowledge Distillation by using the interrelationships between layers as inputs for the student model. These relationships include layers that function as probability distributions, similarity matrices, or feature embeddings. The student model learns from the teacher model on how to construct its own embeddings or distributions based on these interrelationships. This approach enables the student model to effectively leverage the knowledge of the teacher model for improved performance.

\section{Methodology}
First, we applied the Segment Anything Model (SAM) on unlabeled thermal infrared images to generate segmentation results. SAM can obtain masks with two types of prompts: points and regions. For the generalization ability of our method in various fields, we used the entire slide of the image as the input prompt without any special settings. SAM generated segmented masks for each object in the input image as shown in Fig. \ref{framework}(a). The masks provided are of high prediction quality above a certain threshold, indicating that the output masks represent the most salient or representative objects in the corresponding scenes. Consequently, we utilized these masks to generate pseudo labels for our dataset.
\subsection{Generating Pseudo Labels from The Masks}
\label{makelabels}
		\begin{figure}[htbp]
		\centering
		\includegraphics[width=3in]{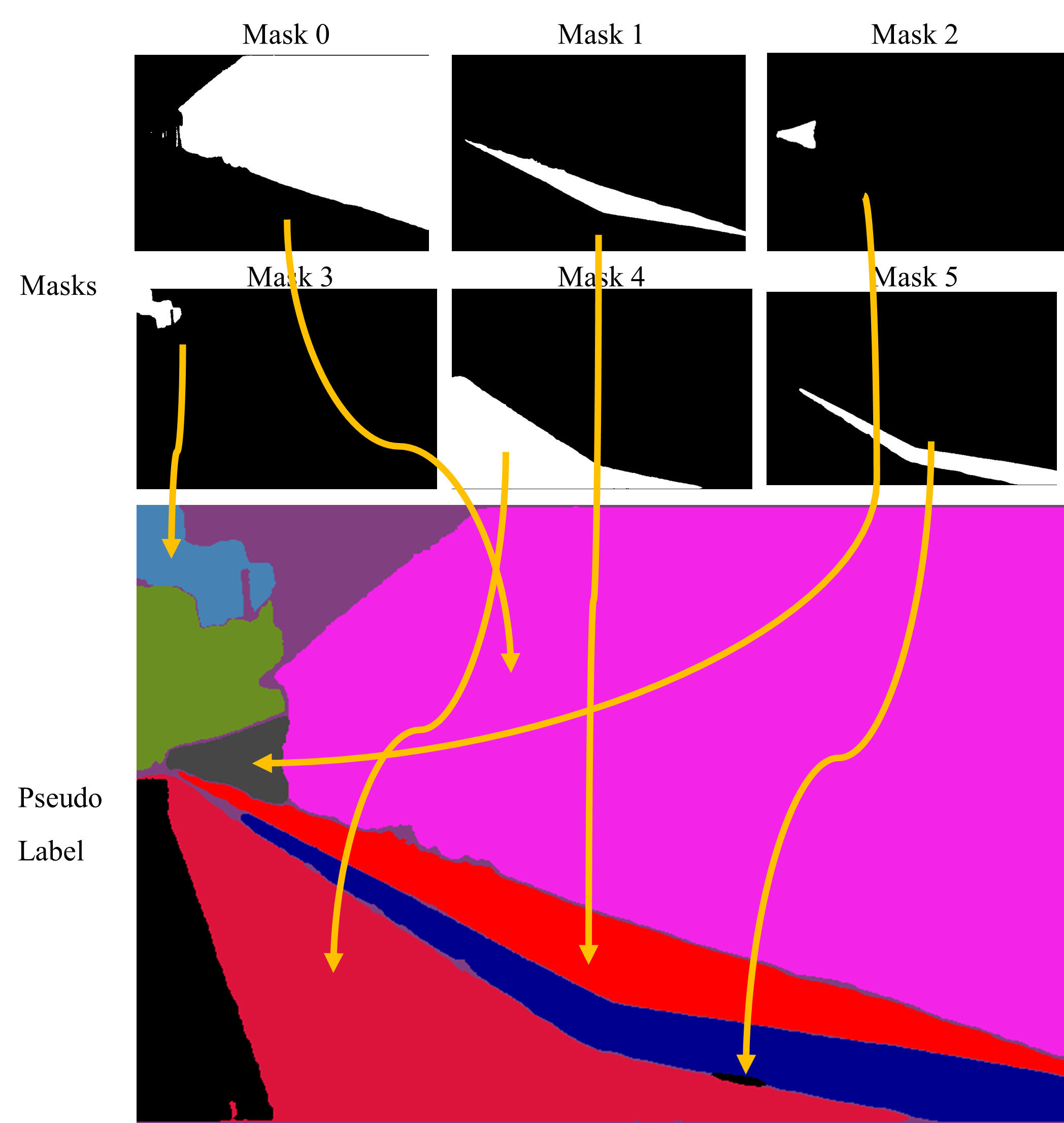}
		\caption{Illustration of the process of generating pseudo labels from masks provided by the Segment Anything Model (SAM). The masks are given in a descending order of quality, and the first few masks are used to build the pseudo labels. The category of each mask is given by its corresponding number, which is used in the final pseudo label.}
		\label{mask2label}
	\end{figure}
The generated masks are ordered in a descending order of quality. Therefore, we can use the top-ranked masks to construct pseudo labels. Each mask is assigned a category number based on its position in the list. For instance, the seventh mask is labeled as category 7 in the final pseudo label. Rest of the images is set to be the background. The process is illustrated in Fig. \ref{mask2label}.

\subsection{Obtaining the Pretrained Model}
After generating the pseudo labels, the resulting dataset can be treated as a semantic segmentation dataset with categories assigned based on the mask index. To extract features and obtain a pretrained model, we use a method of knowledge distillation by training a model on this dataset. After training on this dataset, the resulting model can be extracted as a pretrained model for finetuning on the actual dataset as shown in Fig. \ref{framework} (b) and (c).

\section{Segment Anything in Thermal InfraRed Images Dataset (SATIR)}
		\begin{figure*}[htbp]
		\centering
		\includegraphics[width=7in]{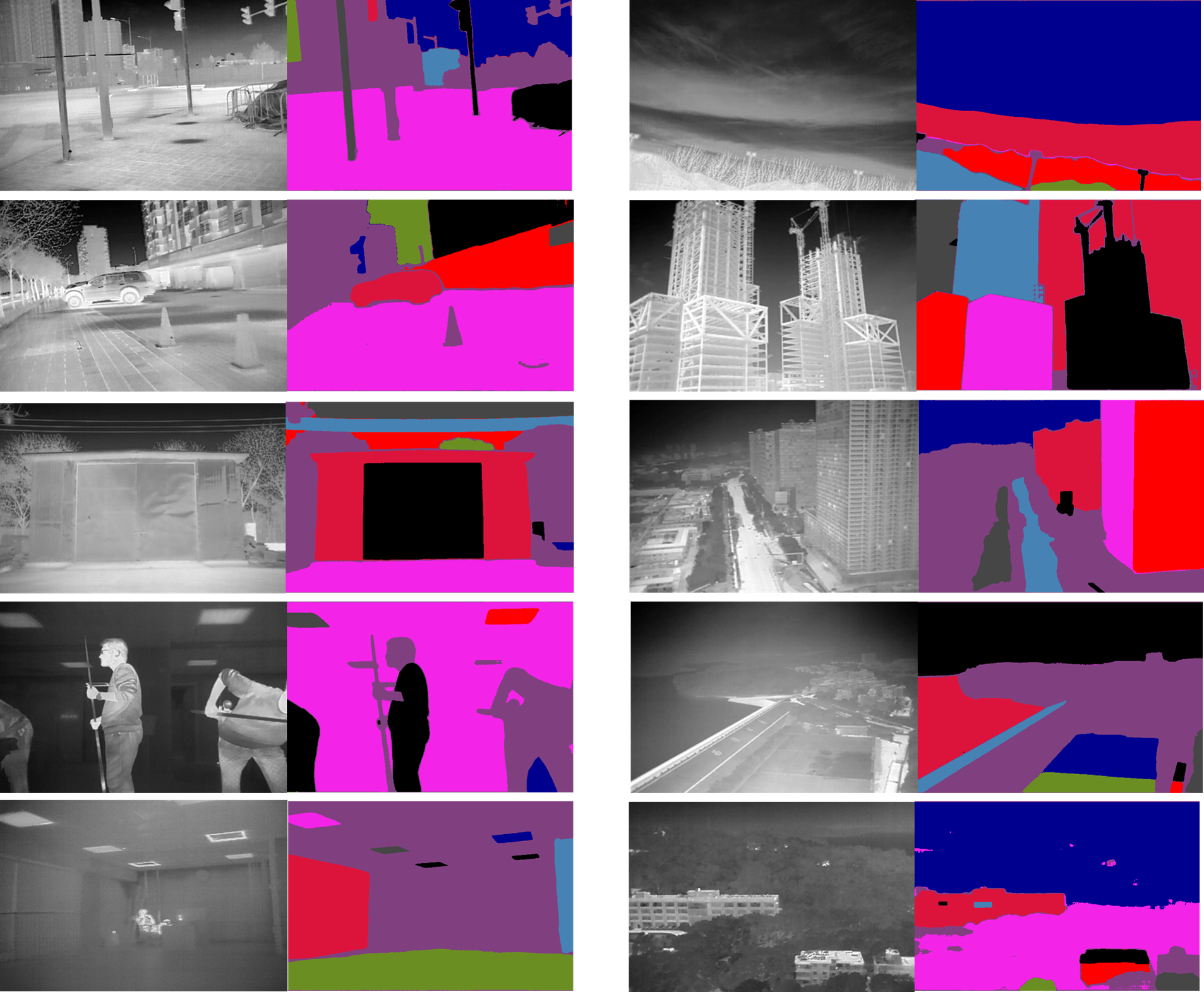}
		\caption{Samples of the SATIR dataset, a large scale thermal infrared dataset with pixel level annotation pseudo labels. The dataset consists of various scenes such as urban, indoor, outdoor, and aerial.}
		\label{datsset}
	\end{figure*}
We collected a large-scale thermal infrared dataset named SATIR, which includes pixel-level annotation labels, as part of our framework. The dataset consists of over 100,000 unlabeled thermal infrared images, and we generated the corresponding labels using the approach described in Section \ref{makelabels}. The images in SATIR cover a wide range of scenes, including urban, indoor, outdoor, and aerial environments, among others. Selected sample images from the dataset are presented in Fig \ref{datsset}.

\section{Experiment}
We evaluated the effectiveness of our proposed framework on the public thermal infrared semantic segmentation dataset, SODA \cite{li2020segmenting}. The SODA dataset contains 2168 real and 5000 synthetically generated thermal images with 21 categories. The real subset of SODA is captured by a FLIR camera (SC260), while the synthetic subset consists of thermal images generated from annotated RGB images. In our evaluation, we used the real subset to assess the performance of our framework. Here we use SegFormer as the backbone and compared our method with no pretrain and ImageNet \cite{deng2009imagenet} pretrained model. In our experiment, we trained all three settings for 240,000 epochs and selected the best performing result.

 \begin{table}[htbp]
 		\renewcommand{\arraystretch}{1.3}
 		\caption{Evaluation results on the SODA dataset. The results show the performance of our proposed framework in terms of Mean Intersection over Union (mIoU) and weighted F-measure ($F_{\beta }^{\omega }$) with no pretrain and ImageNet pretrained model.}
 		\label{tableExp}
 		\centering
 		\begin{tabular}{c|c|ccccc|c}
 			\hline	
            Pretrained with & Backbone model & mIoU $\uparrow$  & $F_{\beta }^{\omega } \uparrow$ \\
            \hline	
   			- & SegFormer & 0.6514 &  0.8156\\
            \hline	
   			ImageNet\cite{deng2009imagenet}& SegFormer & 0.6775 & 0.8374\\
 			\hline
 			SATIR & SegFormer & \bfseries0.6906 & \bfseries0.8426\\
 			\hline
 		\end{tabular}
 	\end{table}

Table \ref{tableExp} presents a quantitative comparison of our method with no pretrain and ImageNet pretrained model on the SODA dataset, where we evaluate the mIoU and $F_{\beta }^{\omega }$ metrics \cite{margolin2014evaluate} among all categories. Our method outperforms the two approaches by approximately 1.3\% in mIoU and 0.5\% in $F_{\beta }^{\omega }$. The experiment demonstrates the capacity of our method for improving the performance of thermal semantic segmentation by pretraining using the ability of SAM.

\section{Limitations and Future Work}
While the proposed model demonstrates the effectiveness of pretraining models on specialized tasks using SAM, there are some limitations that can be addressed in future work.
One major limitation of our proposed method is the crude approach used for assigning labels to the masks. The mask index is currently used as the label, but there are instances of multiple semantic masks with different labels in various annotations. This can potentially lower the performance of the pretrained model on our generated dataset. Future work could explore more sophisticated labeling methods to improve the quality of the generated dataset.

Furthermore, our method presents a paradigm of collaborating with large-scale models trained on vast datasets, which can be extended to various fields facing similar challenges.

\section{Conclusion}
In this paper, we have presented a novel framework for pretraining models for specialized tasks using the Segment Anything Model (SAM) and pseudo labeling. Our approach offers an effective solution for working with special fields where label annotation is challenging, as demonstrated in our experiments with thermal infrared image segmentation. We have shown that our method can effectively improve the accuracy of segmentation results of specific categories beyond the SOTA ImageNet pretrained model. Additionally, we have generated a large scale thermal infrared segmentation dataset used for pretraining, which contains over 100,000 images with pixel-annotation labels. Our proposed framework can also be seen as a paradigm of cooperating with large models trained with large datasets, which can be applied to other fields facing similar challenges. Despite some limitations, such as the rough way of assigning labels for masks, our results show the potential of our approach in advancing pretraining for specialized tasks. We believe our work can contribute to the development of more efficient and accurate models for specialized fields.

\end{document}